# A Thin Format Vision-Based Tactile Sensor with A Micro Lens Array (MLA)

Xia Chen, Guanlan Zhang, Michael Yu Wang, *Fellow, IEEE* and Hongyu Yu, *Member, IEEE*

*Abstract*—Vision-based tactile sensors have been widely studied in the robotics field for high spatial resolution and compatibility with machine learning algorithms. However, the currently employed sensor's imaging system is bulky limiting its further application. Here we present a micro lens array (MLA) based vison system to achieve a low thickness format of the sensor package with high tactile sensing performance. Multiple micromachined micro lens units cover the whole elastic touching layer and provide a stitched clear tactile image, enabling high spatial resolution with a thin thickness of 5 mm. The thermal reflow and soft lithography method ensure the uniform spherical profile and smooth surface of micro lenses. Both optical and mechanical characterization demonstrated the sensor's stable imaging and excellent tactile sensing, enabling precise 3D tactile information, such as displacement mapping and force distribution with an ultra-compact thin structure.

## I. INTRODUCTION

Vision-based tactile sensors have emerged as a new type of tactile sensing method to provide abundant contact information[1-3], such as surface texture[1], depth[2], force distribution[3], contact geometry[4] and slip[5]. High spatial resolution and compatibility with sophisticated image-based machine learning are two significant advantages of vision-based sensors compared to conventional tactile sensors that depend on other transduction principles, such as piezoresistive and capacitive mechanisms. Conventionally, vision-based tactile sensors capture the image using one monocular camera module[1-3]. However, the traditional monocular lens system lacks the miniaturized size and large field of view. Meanwhile, in some other optics applications, prototypes with multiple vision units have been developed, such as TOMBO[6], APCO[7], Cley[8], with compact structures and wide field of views (FOVs). Therefore, in this paper, the microfabricated multiple vision units are adopted as the imaging system for vision-based tactile sensors, and the MLA is utilized in the device for enhancing image quality. Simultaneously, the comparison between pinhole and MLA based on the vision-based tactile sensors is reported.

As the imaging system's key component, the MLAs have been widely applied in optoelectronic systems[9, 10], microelectromechanical systems (MEMS)[11, 12], and bionic devices[13-18]. The lens array's refractive index, dimensions, and uniformity are representative factors for MLAs' characterizations. Compared with micro pinhole arrays, higher resolution and less light intensity requirement are two advantages of MLAs. However, the utilization of MLAs also brings the challenge of optimizing optical distance for both image plane and object plane. Besides, the difficulties in controlling lenses' profiles still exist for the applications requiring a sub-millimeter MLA. In addition, the crosstalk between vision units affects the imaging quality of multi-vision systems as presented in the previous artificial compound eye systems[13, 15, 18]. In our work, the imaging quality between two adjacent vision units is guaranteed by using the proposed walls to isolate the individual lens of MLAs.

Fabricating MLAs is another challenge, since low processing cost, fine control of lens parameters, and smooth surface need to be achieved simultaneously. Among various techniques to develop MLAs, the thermal reflow method[19-21] using lithography can provide a better controllable profile than inkjet printing[22, 23] and hot embossing[24]. Meanwhile, for the lens replication, soft lithography technology is one of the most effective techniques[25]. Among different materials which can be used in the soft lithography technique, polydimethylsiloxane (PDMS) is regarded excellent for its high transmittance[26] and smooth surface[27]. Furthermore, the lens using PDMS can be applied to tunable imaging[18, 28-30] and vari-focus imaging fields[31, 32].

In this study, an MLA enhanced vision-based tactile sensor is developed. Scaled MLAs are designed, fabricated, and assembled in the tactile sensor to satisfy application requirements. The optical properties, such as the focal length and imaging resolution, are quantitively evaluated. The spatial resolution of MLAs is tested compared with the results using pinhole arrays, where the spatial resolution is improved by using MLAs with magnified images. After assembling the imaging components into the tactile sensor, normal and tangential force calibration with commercial force sensor and several objects' contact experiments are executed and evaluated. The sensor's stable image output and response to the contact force (normal and tangential force) verify the excellent sensing abilities of the tactile sensor with MLAs.

*Research supported by grants from the Innovation and Technology Commission (project: ITS/104/19FP) of HKSAR, the Foshan The Hong Kong University of Science and Technology (HKUST) Projects under Grant FSUST19- FYTRI05, and the startup fund from the Hong Kong University of Science and Technology (*corresponding author: Xia Chen*).

Xia Chen and Guanlan Zhang is with the Department of Mechanical and Aerospace Engineering, Hong Kong University of Science and Technology, Hong Kong (e-mail: xchendj@connect.ust.hk; gzhangaq@connect.ust.hk).

Michael Yu Wang is with the Department of Mechanical and Aerospace Engineering and the Department of Electronic and Computer Engineering, Hong Kong University of Science and Technology, Hong Kong (e-mail: mywang@ust.hk).

Hongyu Yu is with the Department of Mechanical and Aerospace Engineering, Hong Kong University of Science and Technology, Hong Kong(e-mail: hongyuyu@ust.hk).

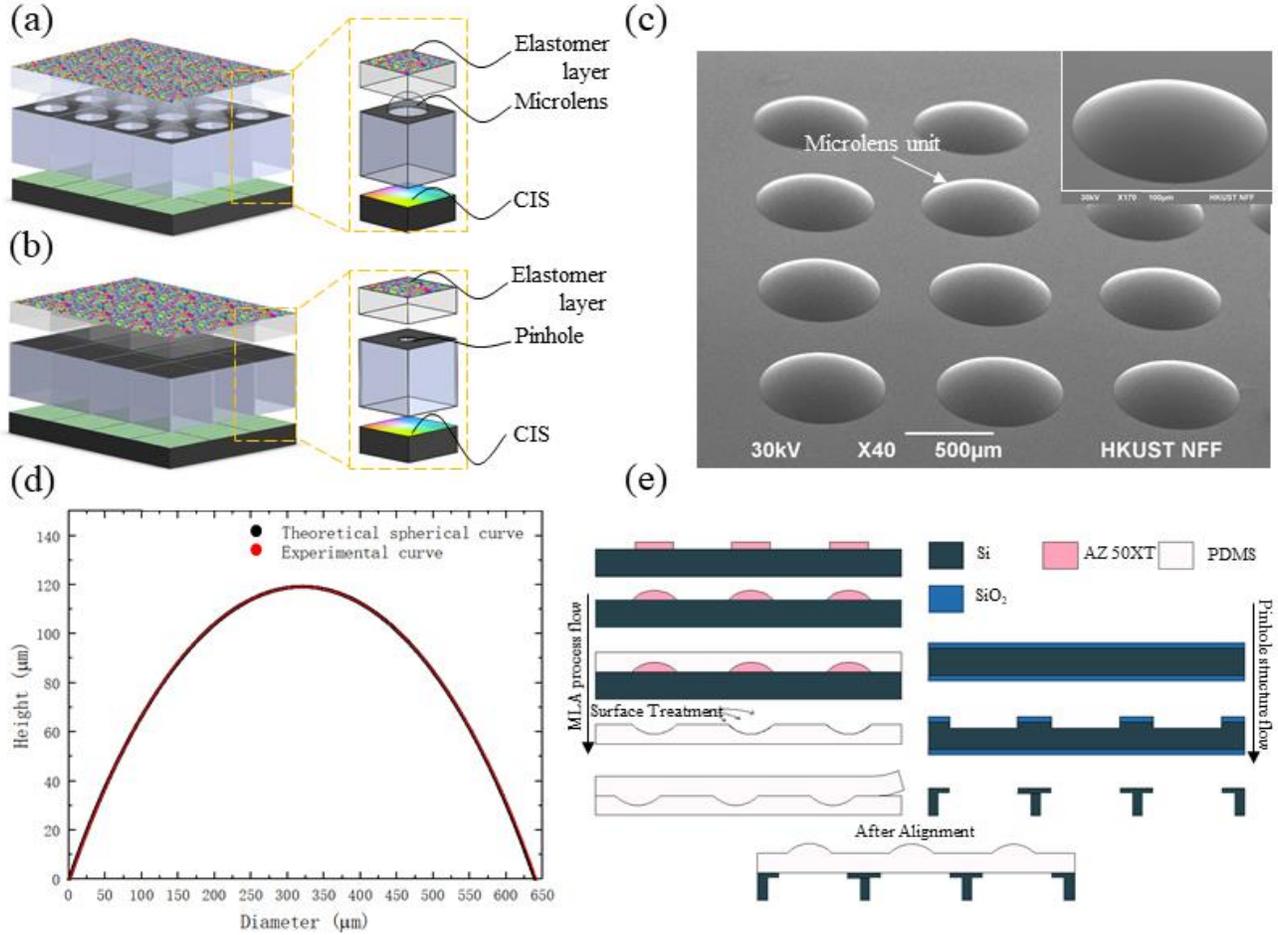

Figure 1. a) Schematic of the multiple vision units based tactile sensor with an MLA b) Schematic of the multiple vision units based tactile sensor with pinholes. c) Surface morphology of the MLA, the inset image illustrates the zoom-in profile of the single lens unit. d) Profile of a single lens unit with a measured and theoretical spherical curve. e) Process flow of fabricating and assembling the MLA on the pinhole structure.

## II. RESULTS AND DISCUSSION

### A. Fabrication of Micro lens array

The schematic of the vision-based tactile sensor using an MLA is shown in Figure 1a. The tactile sensor comprises three main sub-systems: touching, imaging, and supporting sub-systems. The touching sub-system includes an elastomer layer, a pattern layer (where markers are embedded), a reflective layer and a protection layer. Below the touching sub-system, the imaging sub-system consists of the insolated square chamber structure, an MLA and a complementary metal oxide semiconductor (CMOS) image sensor (CIS). The supporting sub-system contains the supporting frame and illumination source. The tactile sensor using the pinhole imaging method is shown in Figure 1b. Compared with the pinhole method, an MLA is placed onto the pinhole structure where the pinhole is used to define the aperture. The combination of MLA and the aperture of pinhole structure can have a more extended depth of field, which benefits the touching sub-system when the elastomer layer is deformed due to external contact. The MLA has a convex shape and faces the touching layer rather than facing the CIS to amplify the imaging details.

The fabricated lenses have excellent smoothness and near-ideal shape in the Figure 1c. The roughness is illustrated by the mean value Ra of 0.024 μm in 1 μm x 1 μm area, which is comparable to previous work[33]. Using the thermal reflow technique, the spherical lens shape is guaranteed, and the measured profile is fits well with the theoretical spherical one (Figure 1d). The Pearson correlation coefficient between these two curves is 0.99994, verifying the strong fitness of the experimental and theoretical results. The spherical curve equation is listed as below,

$$R = \frac{h}{2} + \frac{D^2}{8h} \quad (1)$$

where $R$ is the radius of the spherical shape, $h$ is the sag height and $D$ is the sag diameter. The fabrication process flow of the MLA is demonstrated in Figure 1e. The cylindrical shape was formed after the lithography process, starting from the photoresist AZ 50 XT on the silicon wafer. After reflow in an oven environment, the spherical shape of the photoresist was well formed. PDMS was then poured on top of the reflowed photoresist mold and cured and peeled off as a concave mold. Then the concave mold was treated with trichloro(1H,1H,2H,2H-perfluorooctyl)silane to serve as reusable transferring mold[34]. The plano-convex multiple lens array was finally formed by pouring, curing and peeling off PDMS from the transferring mold. It is worth noting that the utilization of the chlorosilane is necessary for peeling off the plano-convex lens from the concave lens mold. At the same time, the pinhole structure and square isolating chamber structure were formed using double side etching of a silicon wafer[35]. The multiple vision imaging system using MLA on the square isolating chamber structure was aligned and assembled. Here, the sag height of lens was mainly controlled by the process parameters, which are the coating times and spin speed of the photoresist. For example, the height of 120 μm was obtained after triple times coating with 1,000 rotations per minute (rpm) spin speed (Details can be found in the experimental section). The transparency test shows its high transmittance ability (around 95%) in the visible wavelength area, which is close to a commercialized BK7 glass lens.

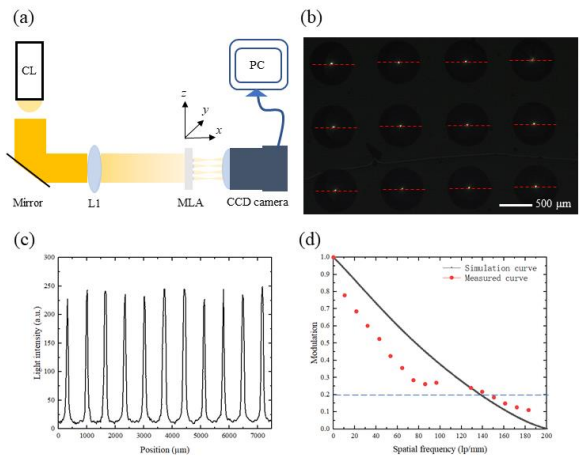

Figure 2 a) Schematic optical properties measurement set up where CL is the collimated light, L1 is the collimator lens. b) Foci distribution of twelve vision units. The red dash line illustrates the lighting intensity testing path. c) Light intensity distribution of an MLA. d) MTF curves of the MLA by calculated spatial frequency response and simulation.

### B. Optical properties

Different to other tactile sensors that transduce contact signals to the electrical output, vision-based tactile sensors rely on images to analyze the force stimuli from outside. To quantitively evaluate the images, the optical properties of MLAs were studied. The MLAs with sag height from 80 μm to 140 μm and different focal lengths and numerical aperture were also tested. After balancing the radius curvature, the pinhole diaphragm radius and the working range of the touching layer (Details in Optical Parameters, Supporting Infomation), the MLA with 640 μm diameter and a 120 μm sag height was selected. With a specific focal length and an image distance, the object distance is determined and at the same time, the depth of field is determined by both the lens and aperture's diameter.

The measurement schematic setup is shown in the Figure 2a. The MLA is placed on the three-axis movement platform. The light is emitted from the lamp and refracted by the mirror. With the collimator lens L1, the light is evenly transmitted to the MLA. In Figure 2b, focusing images of 12 vision units are shown by above experimental setup. The light intensity distribution is shown in Figure 2c. It should be noted that all twelve units demonstrate similar light intensity where the variation is no more than 10%. Figure 2d shows the MLA modulation transfer function (MTF) curve using the slanted-edge calculation from spatial frequency response. The ISO12233 testing chart is used as the edge image source. Using the conventional criterion of 0.2, around 140 lp/mm is achieved by the experimental result which is similar to the simulation curve. Besides, the spatial resolution chart of USAF 1951 is also used for rough testing. The pattern in the red frame clearly shows that the resolution reaches up to 144 lp/mm, which meets the above result.

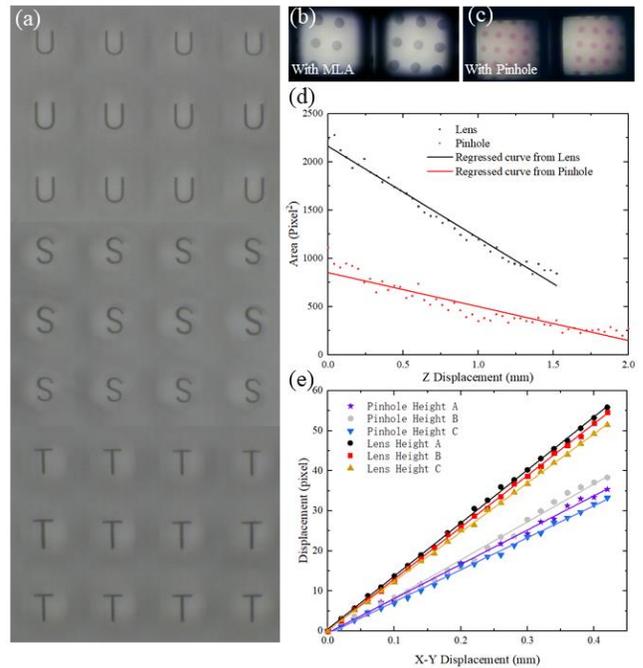

Figure 3 a) Image demonstration using U, S, T letters with an MLA. b) The circular dot image with the MLA. c) The circular dot image with pinhole. d) Area of a circular dot moving in the Z direction with a micro lens array based sample and a pinhole based sample. e) Displacement with circular dot moving in the X-Y direction with a micro lens array based sample and a pinhole based sample.

### C. Imaging test

The imaging test is performed as shown in Figure 3a using the letter U, S and T. To test the spatial resolution used in the tactile sensor, a dot image test was performed using circular dots in Figure 3b and c. The Movie S1 clearly shows the movement with these two imaging methods using the same object. Blob detection is used here to capture the area and the central point displacement. With Z directional moving to the far side, the area of the dot decreases. In Figure 3d, the linearity of the MLA result ($R^2 = 0.98$) is better than the pinhole result ($R^2 = 0.88$). It proves that using MLA leads to a more precise image while pinhole imaging leads to a blurry image. The gradient of the MLA curve is around 2.7 times of the result of the pinhole, indicating the superior ability in the spatial resolution range with magnified images. The X,Y directional movement experiments were carried out as shown

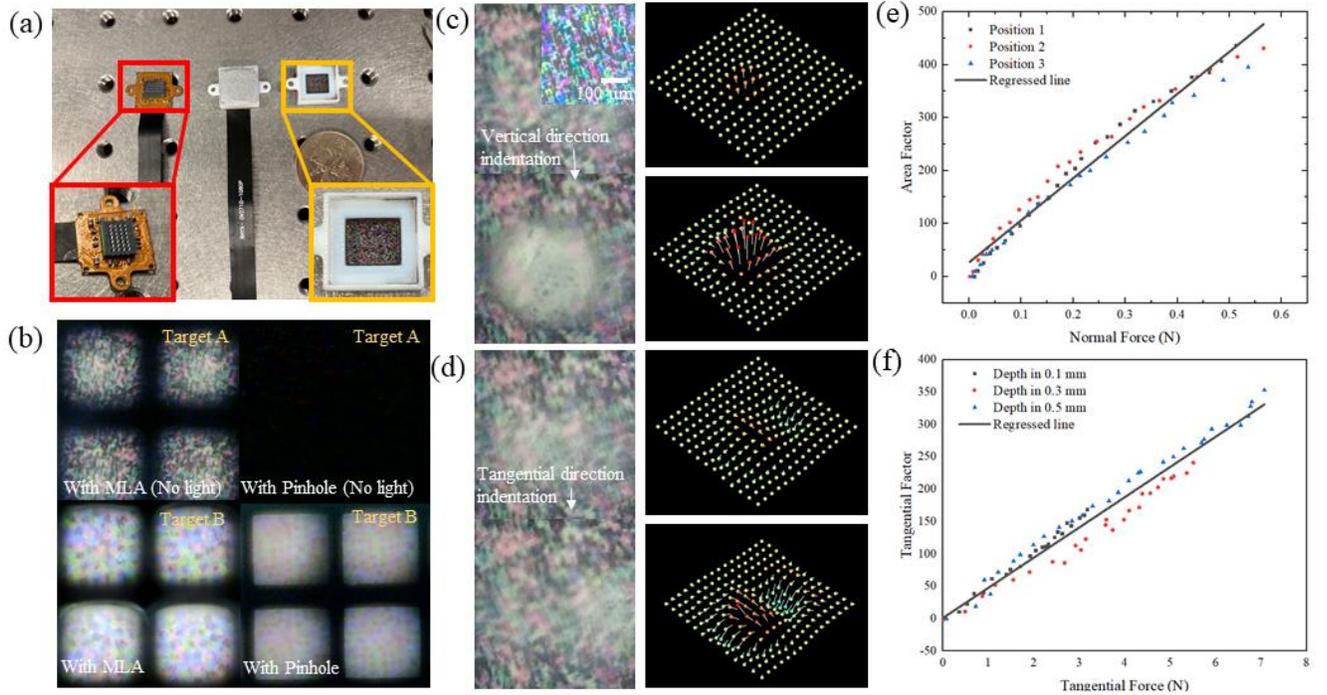

Figure 4 a) Prototype of the tactile sensor with an MLA. The red inset image shows the MLA on the square isolation chamber structure, and the pinhole structure is put on the CIS. The middle one shows the tactile sensor with MLA embedded. The yellow inset image shows the colorful pattern embedded in the supporting frame. b) Two types of color patterns are tested as Target A and Target B. The top two images show the denser color pattern (Target A) using MLA and pinhole without an illumination source, respectively. The bottom two images show the sparser color pattern (Target B) using MLA and pinhole with illumination. c) The left images are the stitched results of normal directional indentation with a spherical object. The inset image illustrates the random color pattern's appearance with a 40X microscope. The deformation fields are demonstrated with a small and a large indentation, respectively. d) The left images are the stitched results of tangential directional indentation with a spherical object. The deformation fields are illustrated with the tangential movement. e) The normal indentation experiment is carried out by pressing three positions on the elastomer layer. The normal force is obtained from the commercialized force sensor while the Area Factor is calculated from the deformation field. f) The tangential movement experiment is carried out by moving the object while keeping the indentation depth in three different levels. The tangential force is obtained from the commercialized force sensor while the Tangential Factor is calculated from the deformation field.

in Figure 3e, and three heights in the working range were adopted (noted as height A, B, C, where height A denotes the 4.8 mm distance between the dot and the CIS, height B denotes 4.9 mm and height C denotes 5 mm). It is worth noting that the gradient of lens curves is higher than the pinhole curves, indicating the higher resolution (around 1.48 times) in the X,Y movement directions. Besides, the nearer the pattern is, the more displacement should occur with the same step length, which means the gradient of height A should be the highest and the gradient of height C should be lowest. The result from the lens agrees well with the above theory, while the pinhole curve result does not fit well. In the results for pinhole curves, the gradient of height A (85.6) is lower than heigh B (94.7), which means that in the small range, the threshold set in the blob detection cannot detect the difference of the dot's displacement between these two heights. It indeed indicates the lower spatial resolution in this working range of pinhole compared with MLA.

### III. STITCHED IMAGE AND FORCE INDENTATION CALIBRATION

The aforementioned imaging sub-system was assembled with the other two sub-systems (Touching sub-system and supporting sub-system) to form the proposed tactile sensors, as shown in Figure 4a. In Figure 4b, two types of color patterns[36, 37] with different pattern densities (Target A and Target B) are adopted as objects using MLA and pinhole, separately. It proves that the light intensity required for MLA imaging is much lower than that with pinhole imaging system. Besides, using MLA in the prototype, a more detailed and magnified image can be obtained with the below images. The MLA contributes to capturing an image without illumination sources when the same image is too dark to recognize using pinoles. In Figure 4c-d, a spherical object is used to press the touching sub-system and two images are illustrated for its deformation procedure. The inset image demonstrates the colorful pattern with a 40x microscope. The Movie S2 demonstrates the normal and tangential movement after stitching the multiple images.

Here we adopt two calculated signals[35], Area Factor (*AF*) and Tangential Factor (*TF*), to calibrate with the reference commercial sensor's normal force and tangential force. The equations to obtain *AF* and *TF* are shown as below,

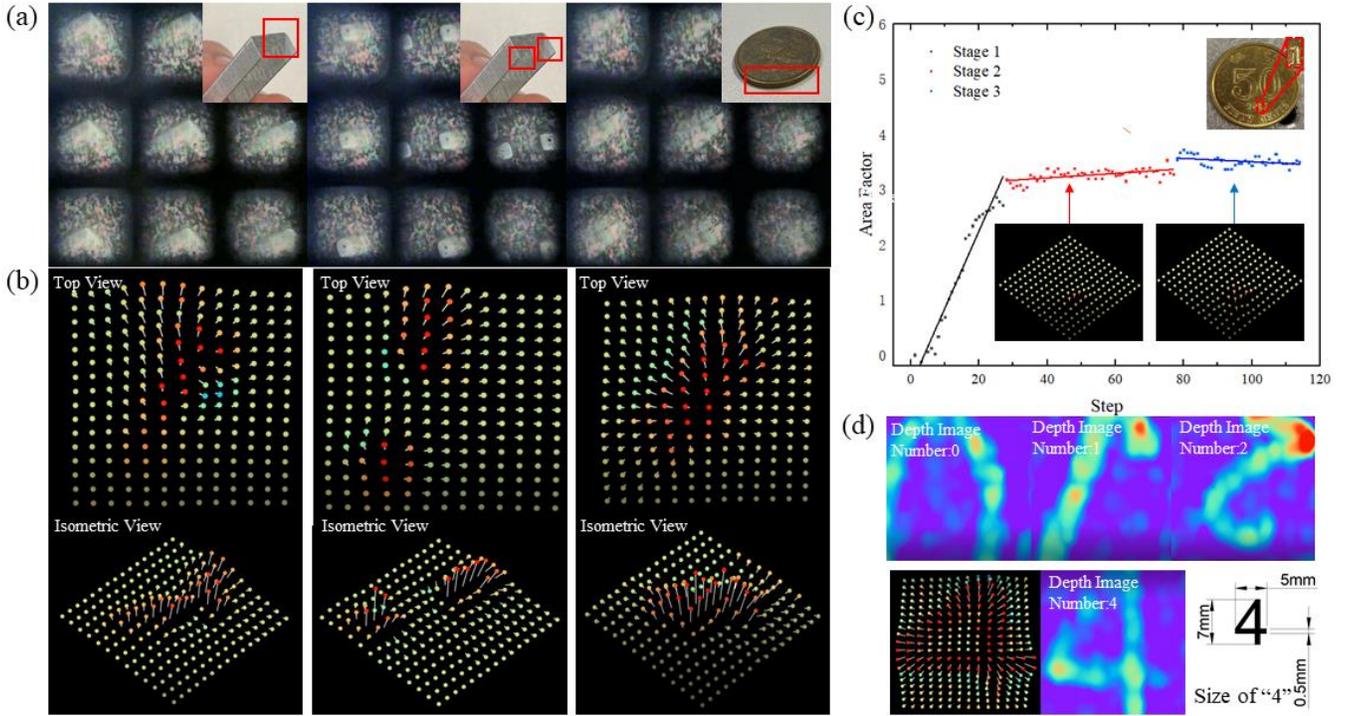

Figure 5 a) Objects' indentation on the tactile sensor. The image on the left is the stapler indentation using its corner. The width of the stapler is around 0.5 mm. The middle image shows the indentation using two sharp points of the stapler. The image on the right shows the rotational movement using the edge of a coin. The raw images before processing clearly show the indentation procedure with small objects. b) Deformation field in top view and isometric view of above three type indentations. Three videos show the indentation process and deformation fields (Movie S3-5). c) Number "1" profile pressing by around 100 μm from stage 2 to stage 3. (d) Depth images of the number "0", "1", "2" which are obtained from the deformation filed images by interpolation. The deformation field, depth image and size of number "4" are listed at the bottom.

$$AF = \sum_{i=1}^{N_p} (\frac{S_i}{S_i^{init}} - 1) \quad (2)$$

$$TF = \left\| \sum_{i=1}^{w \cdot g \cdot h} U_s^i \right\| \quad (3)$$

where $N_p$ is the number of patches segmented for area calculation, $S_i$ and $S_i^{init}$ denote area and area of the first frame of $i^{th}$, $w$ and $h$ are width and height of optical flow frame patch. The computation of $AF$ and $TF$ considers both the accuracy of depth estimation and the average error due to the accumulated pixel tracking error in optical flow. With the increased number of scatter points in AF calculation, the resulting depth estimation can be smoother and closer to the actual surface. However, the result will be less stable due to the oscillating error in the displacement vector field. The computation speed also drops significantly. Therefore, we finally chose 16×13=208 points to achieve a relatively good calculation quality. For the normal indentation (Figure 4e), three positions were tested with the spherical object and the regressed line illustrates the good linearity ($R^2 = 0.97$). For the tangential movement, the depths of 0.1mm, 0.3 mm and 0.5 mm were selected. The deformation images are shown in Figure 4d. Figure 4f gives the calibration result from the calculated TF and tangential force from the sensor. It is noticeable that the deeper the indentation, the bigger the tangential force it will cause. The regressed line ($R^2 = 0.96$) also demonstrates the good linearity of the sensor response.

After calibration with the force sensor, several small objects (stapler and coin edge) were pressed onto the tactile sensor for demonstration as shown in Figure 5a. By the operations of image stitching and surface deformation tracking[35], the deformation fields are obtained in the top view as well as the isometric view in Figure 5b. The deformation fields produced by several objects illustrate the sensor's ability to perceive objects of sub-millimeter size. Furthermore, Figure 5c illustrates the sensors' tiny depth difference (100 μm) while the object is pressed onto the touching layer. Between step 2 and step 3, the object is squeezed down 100 μm and the signal of AF clearly shows the gap by two regressed curves. For depth estimation, more objects with a different shape (number 0, 1, 2, 4) were pressed onto the sensor. It further proves the sensor's stable tactile feedback especially for dimensions in sub-millimeters.

## IV. CONCLUSION

In conclusion, a micro lens array (MLA) enhanced vision-based tactile sensor was fabricated, characterized and tested. The thermal reflow method ensures the controllable profile of the lens compared to other lens forming methods. Unlike traditional micro lens in several microns height, the heights (around 120 μm) and diameters (from 50 μm to 700 μm) of MLAs were produced using the microfabrication approach. The fabricated MLA shows high uniformity and high spatial resolution which reaches up to 140 lp/mm. The optical test shows the excellent imaging properties of the MLAs.

Compared with pinhole imaging, the MLA enables the tactile sensor to have higher spatial resolution and magnified object details. The force calibration further completes the tactile sensor's characterization. The force signals calculated from the deformation field verify the force sensing abilities in reference to a commercial force sensor. With this work, the vision-based tactile sensors can compete with previous prototypes with pinhole structures concerning spatial resolution and deformation tracking accuracy. More importantly, more advanced MLAs based vision-based tactile sensors, such as the tactile sensor with a wide field of view and vari-focal function, can be developed based on this work.

## APPENDIX

MLA fabrication process: The convex multiple lens array was first formed in a cylinder using AZ 50 XT (Clariant AG). According to the different requirements of the MLA's profile, the photoresist was spin-coated on a silicon wafer several times. Karl Suss MA6 aligner was used for the lithography process with a power density of around 23 mJ/cm$^2$. AZ Developer 400K (AZ Electronic Materials) was utilized for the developing stage. After developing, the photoresist was then placed in an oven at 120 ℃ over 12 hours. Then, PDMS (Sylgard 184 Silicone, Dow) was mixed with a 10:1 ratio of basing material and curing agent. After degassing, the mixture was spin-coated on the silicon wafer. The concave PDMS MLA mold was obtained after 4-hour baking in an 80 ℃ oven. Trichloro (1H,1H,2H,2H-perfluorooctyl) silane (Sigma, Aldrich) was used to treat the surface of concave MLA mold. The second PMDS (mixing ratio of 5:1) molding was then processed to obtain the convex MLA.

Morphology Characterization: The morphology images were captured by the scanning electron microscope (Model JSM-6490, JEOL). The lens' height and diameter were measured by the profile (P-10, KLA-Tencor).

Optical Property Characterization: The transmittance of the PDMS was measured by the UV/VIS spectrophotometer (Lambda 20, Perkin Elmer). The LED ring shape lighting source (MIC-199, POMEAS) was calibrated by the lens (GCO-2410, Daheng Optics). The focusing and letter image tests were obtained by a camera (DFK 33UP5000, Imagingsource) with 1 inch ON CMOS image sensor with 4.8 μm pixel size. The MTF curve was carried out with the help of open-source ImageJ[38]. The simulation result of the MTF curve was obtained using Zemax. Image sensor (OV 2710, OmniVision) was utilized for the tactile sensor device.

## ACKNOWLEDGMENT

This work was supported by grants from the Innovation and Technology Commission (project: ITS/104/19FP) of HKSAR, the Foshan The Hong Kong University of Science and Technology (HKUST) Projects under Grant FSUST19-FYTRI05, and the startup fund from the Hong Kong University of Science and Technology. We acknowledge the Nanosystem Fabrication Facility (NFF) of HKUST for the device/system fabrication. We appreciate Dr. Mark Ellwood for proofreading and English editing.